\documentclass[runningheads]{llncs}
\usepackage[T1]{fontenc}
\usepackage{graphicx}
\usepackage{hyperref}
\usepackage{color}
\usepackage{svg}
\usepackage{multirow}
\usepackage{makecell}

\begin{document}

\title{Semantic Latent Space Regression of Diffusion Autoencoders for Vertebral Fracture Grading}
\titlerunning{Semantic Latent Regression of DAE for Vertebral Fracture Grading}
\author{Matthias Keicher\inst{2}\thanks{Equal contribution.}
\and Matan Atad\inst{1}$^*$
\and David Schinz\inst{4}
\and Alexandra S. Gersing\inst{4,5}
\and Sarah C. Foreman\inst{4}
\and Sophia S. Goller\inst{5}
\and Juergen Weissinger\inst{5}
\and Jon Rischewski\inst{5}
\and Anna-Sophia Dietrich\inst{4}
\and Benedikt Wiestler\inst{4}
\and Jan S. Kirschke\inst{4}
\and Nassir Navab\inst{2,3}}
\authorrunning{M. Keicher et al.}
%
\institute{Technical University of Munich, Germany \and
Computer Aided Medical Procedures, Technical University of Munich, Germany \and
Johns Hopkins University, USA \and
Klinikum Rechts der Isar (Technical University of Munich) \and
Klinikum der Universität München (University of Munich)
}
\maketitle              
\begin{abstract}

Vertebral fractures are a consequence of osteoporosis, with significant health implications for affected patients. Unfortunately, grading their severity using CT exams is hard and subjective, motivating automated grading methods. However, current approaches are hindered by imbalance and scarcity of data and a lack of interpretability.
To address these challenges, this paper proposes a novel approach that leverages unlabelled data to train a generative Diffusion Autoencoder (DAE) model as an unsupervised feature extractor. We model fracture grading as a continuous regression, which is more reflective of the smooth progression of fractures.
Specifically, we use a binary, supervised fracture classifier to construct a hyperplane in the DAE's latent space. We then regress the severity of the fracture as a function of the distance to this hyperplane, calibrating the results to the Genant scale. Importantly, the generative nature of our method allows us to visualize different grades of a given vertebra, providing interpretability and insight into the features that contribute to automated grading.

\keywords{Vertebral Fracture Diagnosis \and Interpretability \and Diffusion Model \and Latent Space \and Generative Feature Extractor}

\end{abstract}

\section{Introduction}

Vertebral Compression Fractures (VCFs) are the most prevalent osteoporotic fractures, occurring in 30–50\% of the population over the age of 50 \cite{ballane2017worldwide}. They develop when a bony block or vertebral body in the spine collapses, causing significant pain and inability to perform daily activities \cite{old2004vertebral}. When detected early, osteoporosis can be effectively treated to reduce future fractures and morbidity. However, prior to the outbreak of pain due to fractures, osteoporosis is commonly asymptomatic, leading to underdiagnosis. To determine the severity of a fracture, radiologists examine CT images to infer the Genant scale \cite{genant1993vertebral}, which measures the reduction in vertebra height compared to its baseline. However, the quality of this measurement depends on the radiologist's experience and skill. The scale has four levels: G0 (Normal/No fracture), G1 (Mild/Up to 25\% height reduction), G2 (Medium/25-40\% reduction), and G3 (Severe/Over 40\% reduction).\looseness=-1

Deep Learning can automate the VCF detection process, and indeed, several such approaches were developed in recent years \cite{bar2017compression,valentinitsch2019opportunistic,tomita2018deep,chettrit2020,husseini2020grading,yilmaz2021automated,engstler2022interpretable,windsor2022context}. However, only a few works so far have considered VCF grading, all using fully-supervised settings \cite{pisov2020keypoints,zakharov2022interpretable,wei2022faint}. Compared to fracture detection, grading is an even more imbalanced task since medium to severely fractured vertebrae account for only a small portion of overall data. Moreover, though vertebrae segmentation datasets exist, only a few graded vertebra CTs are available. To address this kind of imbalance in computer-aided diagnosis, generative models have been proposed for data augmentation of underrepresented data \cite{frid2018synthetic,kitchen2017deep,nie2017medical,sankar2021glowin} while another line of work \cite{yi2018unsupervised,atad2022chexplaining,nitzan2021large,preechakul2022diffusion} makes direct use of the latent code of generative models. Recently, \cite{botros2022vertebral} applied an anomaly scoring approach, comparing a vertebra silhouette to a healthy one generated by StyleGAN2 \cite{karras2020analyzing} to derive a fracture grade, but this has outliers, and no evaluation results were given. In this work, we use an unsupervised generative model, the Diffusion Autoencoder (DAE) \cite{preechakul2022diffusion}, to extract features for grading VCFs. We model the grading as a linear regression in the latent space of the DAE and show its semantic expressiveness.

Our contribution is three-fold: (1) \textbf{Unsupervised feature extraction}: we use unlabelled data to train a generative DAE model as an unsupervised feature-extractor. Unlike StyleGAN2-based approaches like LARGE \cite{nitzan2021large} our approach eliminates the need to train an encoder for GAN inversion \cite{tov2021designing}.
(2) \textbf{Modelling fracture grading as a continuous regression}: in contrast to most methods, that use a multiclass setting to model the ordinal regression of Genant fracture grading, we employ a continuous regression approach in the latent space. This modeling is closer to the smooth progression of fractures, yet our method only requires binary fracture labels. (3) \textbf{Inherent interpretability}: the generative nature of DAE allows us to generate images that resemble the model's inner representation of different Genant grades of a given vertebra.  This extends binary counterfactual explanations \cite{atad2022chexplaining} to ordinal regression and aids interpretability by visualizing the change in the image parts that the model deems most important.\looseness=-1

\begin{figure}[t]
  \centering
  \includegraphics[width=1.0\linewidth]{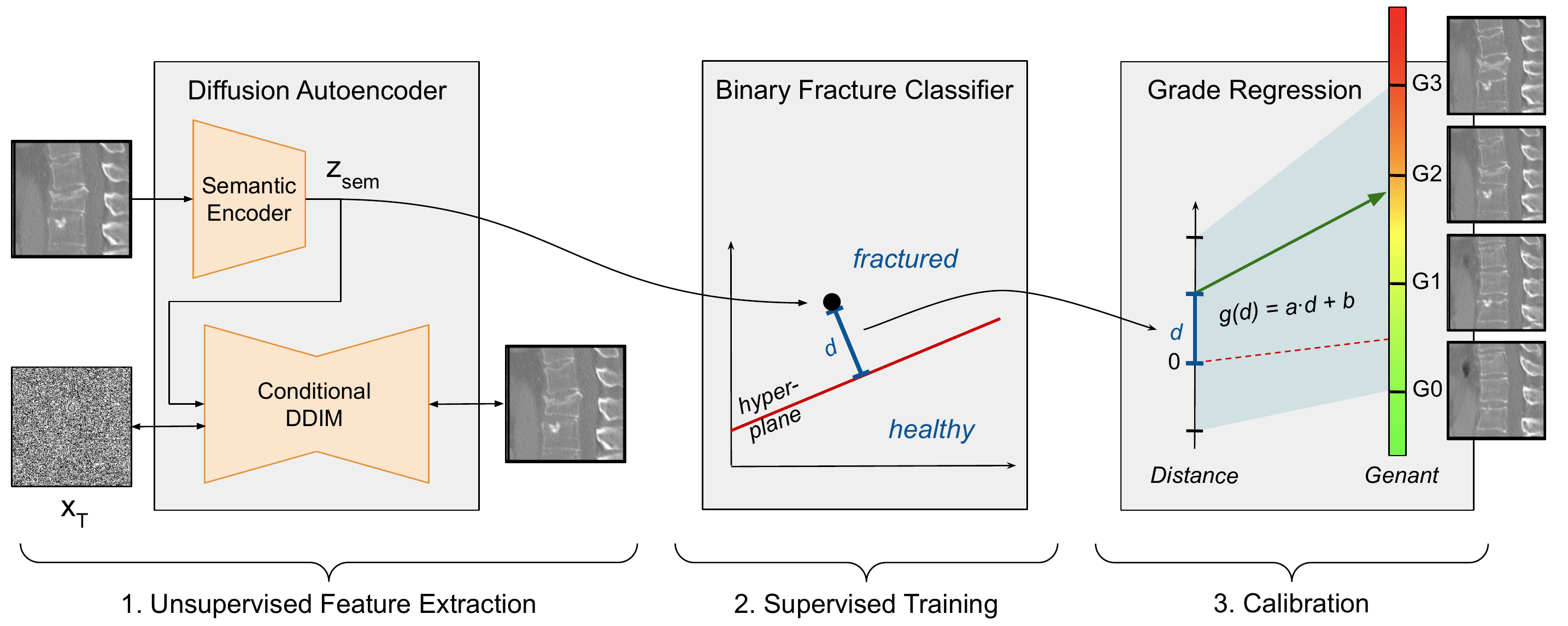}
  \caption{The proposed vertebral fracture grading method involves three steps: 1) unsupervised training of a generative feature extractor using a Diffusion Autoencoder (DAE), 2) supervised training of a binary classifier (G0 vs. G2/3) to detect fractures and obtain a decision hyperplane, and 3) calibrating a linear regression of the Genant grade to the hyperplane distance of embedded images.}
  \label{fig:graphical_abstract}
\end{figure}

\section{Method}

The fracture grading method involves three steps (Fig.~\ref{fig:graphical_abstract}). First, a DAE is trained using unlabelled data. This model learns a compressed, semantically rich latent space useful for downstream tasks. Next, a smaller set of fracture-labeled images is encoded into this latent space, and a linear classifier is trained for binary fracture detection. Finally, the decision boundary of this classifier is extracted as a hyperplane, and a regression model is fitted on the distance of encoded images to the plane to calculate the fracture grade of the vertebral body.

\subsubsection{Diffusion Autoencoder} Diffusion probabilistic Model (DPM) \cite{song2020denoising} and Denoising Diffusion Implicit Model (DDIM) \cite{song2020denoising} are two generative models that learn a denoising process from noisy input to images $x_0$. DPM learns to map noisy inputs to clean images after $T$ denoising passes, whereas DDIM uses a deterministic approach to obtain a single noise latent for a given image. However, both models do not capture high-level semantic information in their latents $x_T$. In contrast, the recently introduced Diffusion Autoencoder (DAE) \cite{preechakul2022diffusion} includes a semantic encoder on top of the DDIM that learns to map input images to semantic meaningful latents $z_{sem}$, useful for downstream tasks. The DDIM is then conditioned on both latents to reconstruct the input image $(x_T, z_{sem}) \rightarrow x_0$, where both the semantic encoder and the DDIM are trained simultaneously.

\subsubsection{Fracture severity regression} \label{sec:regression}
The DAE is trained on a vertebrae dataset without any labels. Next, a linear layer or an SVM classifier is trained to predict the existence of a fracture in the input images based on the semantic latent $z_{sem}$ using a smaller subset of the data with binary labels. Since the labels for grading fractures are very noisy and subjective, only the extreme grades G2 and G3 are used to train the fracture detectors. While the grading of fractures is generally considered a multiclass \cite{wei2022faint} or ordinal regression problem \cite{husseini2020grading}, we intentionally model it as a regression since the compression of osteoporotic vertebrae is a continuous process. For this, the decision boundary of the binary classifiers is extracted. It is represented by the hyperplane $\vec{P}$ given by its normal equation: $\vec{n} \cdot \vec{w} + b$, where $\vec{n}$ is a semantic direction corresponding to the VCF existence, $\vec{w}$ is an input image latent, and $b$ is a bias term. The magnitude of a fracture severity in the image is estimated using the distance of its latent code $\vec{w}_i$ to this hyperplane \cite{nitzan2021large}: ${dist(\vec{w}_i, \vec{P}) = \vec{n} \cdot \vec{w}_i + b}$. To calibrate this distance to the Genant scale, the mean distance for healthy vertebrae and G3 vertebrae is computed and a simple linear regression is fitted. Finally, the continuous values are rounded to the nearest grade to obtain the clinical grading in ordinal categories.

\subsubsection{Semantic image manipulation}
The benefit of the autoencoder architecture is that no separate encoder has to be trained to project input images into the latent space, in contrast to previous works that used StyleGAN (GAN Inversion) \cite{nitzan2021large,xu2021generative}. To generate altered images, the semantic latent code can be changed in the semantic direction $\vec{n}$ and together with the original stochastic latent, decoded by the conditional DDIM to a new image. To generate an image of a particular Genant grade, the calibration utilized in the regression process is inverted to determine the magnitude of the required change along the semantic vector.

\section{Experiments}

\subsubsection{Dataset} We use the public VerSe dataset \cite{liebl2021computed,sekuboyina2021verse} and an in-house dataset acquired at Klinikum Rechts der Isar and Klinikum der Universität München, containing a total of 12019 sagittal 2D CT slices of vertebrae. Each slice has a size of $96 \times 96$ pixels centered around a single vertebra, though multiple surrounding vertebrae are also visible. For each slice, the existence of a VCF in the center vertebra is indicated. Of the 1248 fractured samples, 220 have a Genant grading (74 G1, 102 G2, 44 G3). For training the fracture classifiers and evaluating their grading, we only use samples from the VerSe dataset\footnote{More details about the training setup and datasets can be found under \hyperlink{supplement}{Supplementary material}.\looseness=-1}

\subsubsection{Image encoding and generation}
In the initial set of experiments, we assessed the generative models' capacity to encode previously unseen images in their semantic latent space and subsequently reconstruct them. We measure the perceptual similarity between the original and encoded image using Learned Perceptual Image Patch Similarity (LPIPS) and the general image generation quality using the Fréchet Inception Distance (FID). When training the DAE, images were generated using $1 \times 512$ semantic latent codes, which were concatenated with the "stochastic" latents. We did not apply any image augmentations and kept other training parameters with their defaults. We also experimented with training on a dataset from which the cervical vertebrae (C1-C7) were omitted since medium to severe VCFs in that area of the spine are uncommon \cite{wood2014management}.
To be able to compare our results to StyleGAN2 as a baseline, the state-of-the-art Encoder4Editing (E4E) \cite{tov2021designing} was used.

\subsubsection{Fracture detection and grading}
We trained linear binary classifiers on latent embeddings from the frozen generative models on the \textit{fractured} vs. \textit{healthy} task. Specifically, we use semantic latent codes from the DAE encoder and the encodings of E4E into StyleGAN2's $W^{+}$ latent space, respectively, to train both a simple fully-connected linear layer and a Support Vector Machine (SVM). As a baseline, we trained a DenseNet121 on the input images of the same subset. The linear layer and DenseNet were trained with a cross-entropy loss. The binary detection is evaluated with the area under the receiver operating characteristic curve (AUC) of grade G0 vs. G2/G3. 
For grading performance, we calculated the macro $F_1$ score across G0, G2, and G3. Predicted grades below G2 are considered healthy. For qualitative evaluation, we use the hyperplane to semantically edit borderline cases to assist physicians in diagnosing by \emph{exaggerating} features of subtle fractures to manifest the deterioration potential of mild cases. Finally, we evaluate the calibration of the regression model to the Genant scale by generating a corresponding image for each Genant grade for selected samples.\looseness=-1

\section{Results and discussion}

\begin{table}[t]
  \centering
  \caption{Quantitative evaluation of image encoding and generation. DAE trained without cervical vertebrae levels shows the best performance in encoding and generation.}
  \begin{tabular}{l|l|l|l}
    Model & Encoder & Encoding (LPIPS $\downarrow$) & Generation (FID $\downarrow$) \\
    \hline
    StyleGAN2 & E4E & $0.098$ & $134$ \\
    DAE w/ C1-C7& DAE encoder & $0.069$ & $46$ \\
    DAE & DAE encoder & $\mathbf{0.040}$ & $\mathbf{40}$ \\
  \end{tabular}
  \label{tab:combined}
\end{table}

\begin{table}[t]
  \centering
    \caption{Quantitative evaluation for fracture detection and grading tasks. The linear layer on top of DAE did not converge for grading due to severe data imbalance. Results marked with $^*$ have been evaluated on a different test set and inclusion criteria in 3D. }
      \begin{tabular}{l|l|l|l|l}
        Model & Encoder & Classifier & Detection (AUC $\uparrow$) & Grading ($F_1$ $\uparrow$)  \\
        \hline
        \multicolumn{5}{c}{\textit{Linear probing trained on G0, G2 and G3 with frozen encoder of generative model}}\\
        \hline
        StyleGAN2 & E4E & SVM & $0.74$ & -\\
        DAE & DAE & Linear layer&  0.96 &  (0.23)\\
        DAE& DAE &  SVM & 0.93 &  0.59\\
        \hline
        \multicolumn{5}{c}{\textit{Linear regression with distance to hyperplane, calibrated with means of G0 and G3}}\\
        \hline
        DAE & DAE & Linear layer&  0.96 & 0.44 \\
        DAE& DAE &  SVM & 0.93 &  0.51\\
        \hline
        \multicolumn{5}{c}{\textit{Polynomial regression with distance to hyperplane, calibrated with G0, G2 and G3}}\\
        \hline
        DAE& DAE &  SVM  (deg=1)& 0.93 &  0.42\\
        DAE& DAE &  SVM  (deg=3)& 0.93 &  0.56\\
        \hline
        \multicolumn{5}{c}{\textit{End-to-end training with full supervision  (G0, G2 and G3)}}\\
        \hline
        \multicolumn{3}{l|}{2D DenseNet121, end-to-end baseline} & $0.98$ & $0.65$ \\
        \multicolumn{3}{l|}{3D SE-ResNet50 with SupCon loss \cite{wei2022faint}} & $0.99^\ast$ & $0.86^\ast$ \\
      \end{tabular}
  \label{tab:fracture_results}
\end{table}

\begin{figure}[t]
  \centering
  \includegraphics[width=0.7\linewidth]{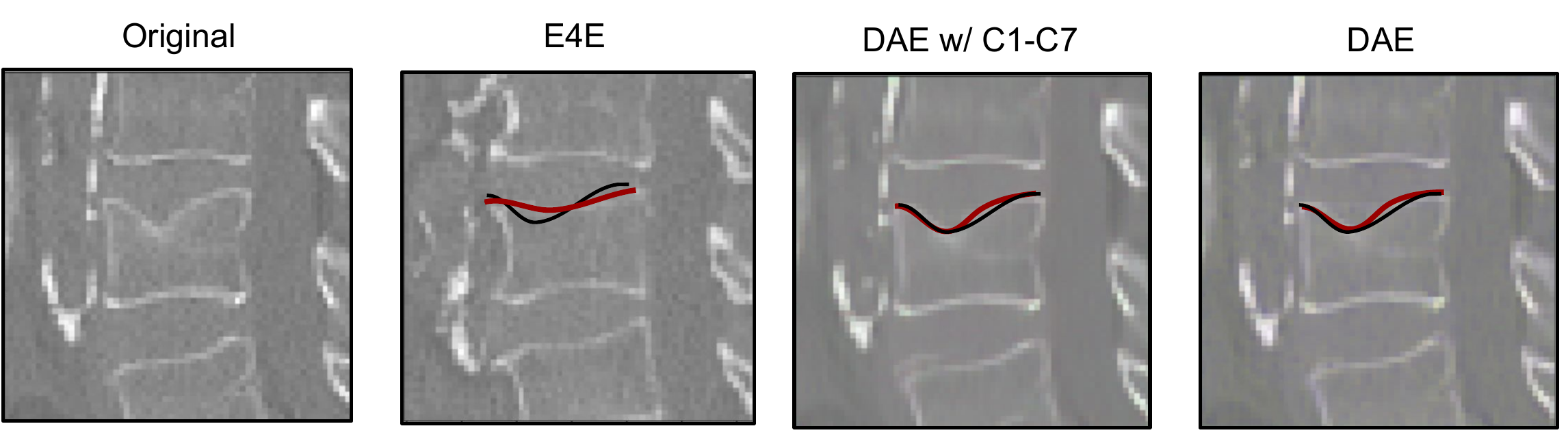}
  \caption{
Qualitative comparison of different encoders: the original shape of the evaluated vertebra is highlighted in black while the reconstructed shape is shown in red. The DAE shows the closest resemblance to the original.
  }
  \label{fig:compare_encoders}
\end{figure}

\begin{figure}[t]
  \includegraphics[width=1\linewidth]{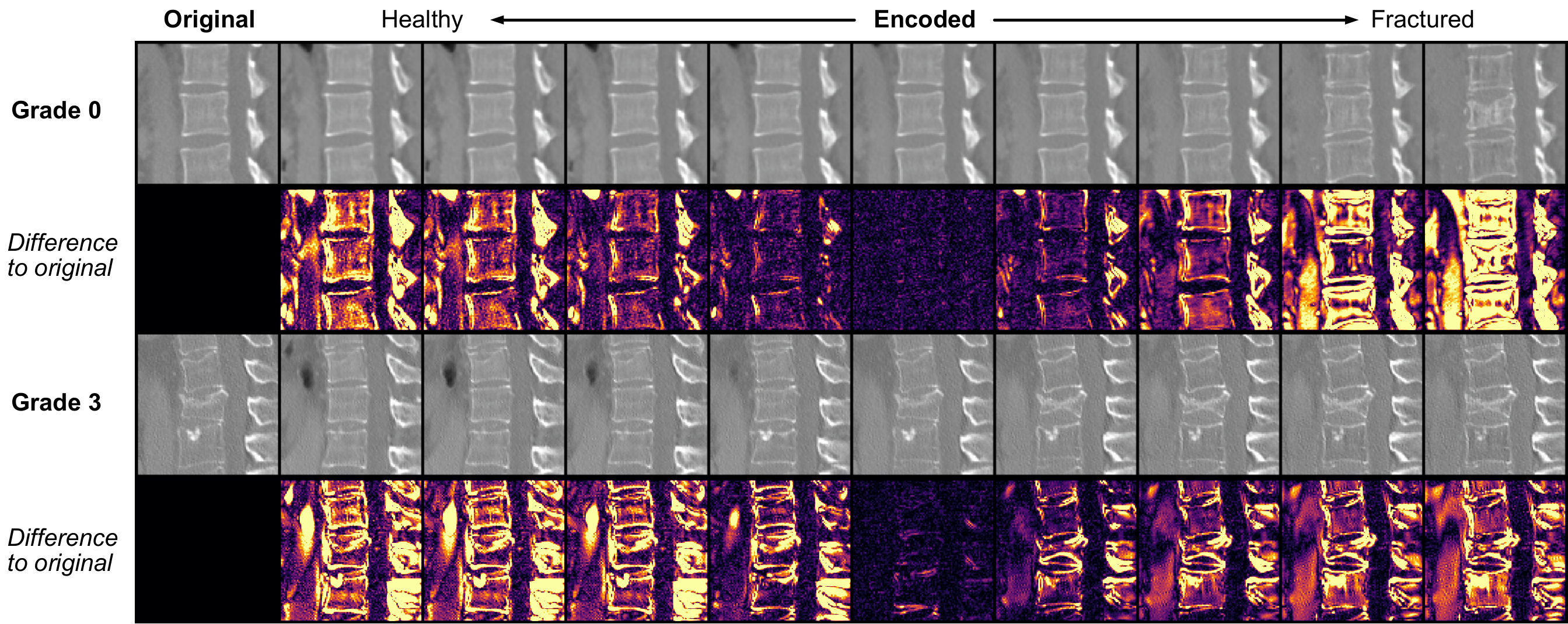}
  \caption{Generated images using our pipeline by moving the semantic latent orthogonal to the hyperplane without calibration. Top row: healthy vertebra (G0) moved in both directions, revealing a severe fracture on the right. Bottom row: severely fractured vertebra (G3) decompresses on the left and further disintegrates on the right.}
  \label{fig:edited_images_shift}
\end{figure}

\begin{figure}
  \includegraphics[width=1\linewidth]{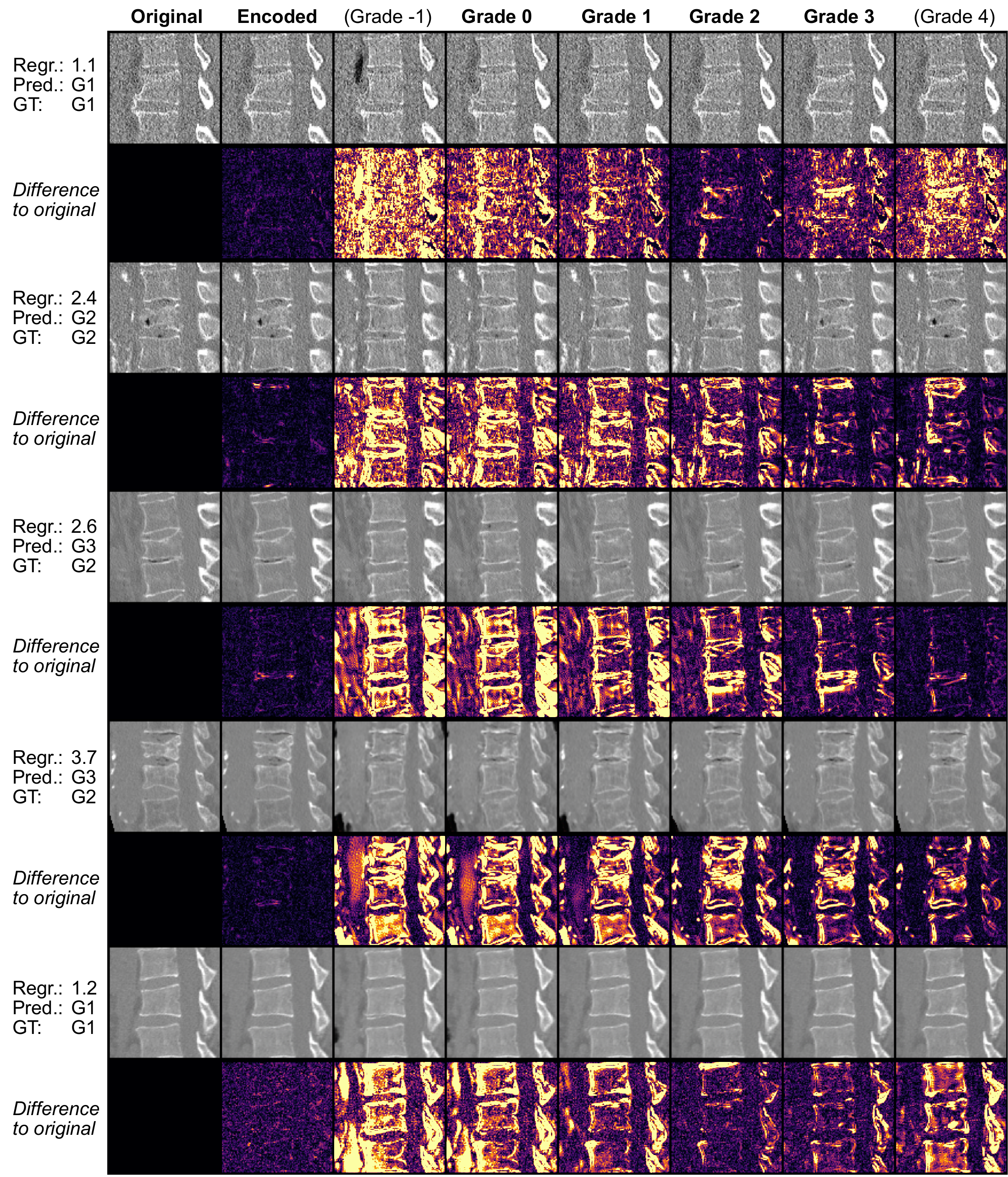}
  \caption{DAE image generation calibrated to Genant Grades (linear regression to SVM hyperplane): On the left, the results of regression, prediction, and the ground truth (GT) are shown. The first three rows are well-calibrated examples, while the bottom two rows show examples that are not well-calibrated. Note that G1 was not used for training the classifiers, and both G1 and G2 were not used for the calibration of the regressors. The artificial scores -1 and 4 are added for illustrational purposes only.}
  \label{fig:edited_images_calibrated}
\end{figure}

Table~\ref{tab:combined} reveals that the DAE semantic encoder is quantitatively superior to E4E in reconstructing vertebrae images based on LPIPS. Fig.~\ref{fig:compare_encoders} confirms this finding by demonstrating the limited ability of E4E to capture fracture-relevant features. Furthermore, the FID measurements in Table~\ref{tab:combined} indicate that DAE generates higher-quality image distributions.
Table~\ref{tab:fracture_results} shows that both linear layers and SVM can effectively separate fractured from healthy vertebrae by constructing a separating hyperplane with a respective AUC of 0.96 and 0.93. While the detection results of the linear layer are better, the hyperplane constructed by SVM shows better results in the linear regression of Genant grades. The qualitative results in Fig.~\ref{fig:edited_images_shift} and \ref{fig:edited_images_calibrated} show that the linear regression model can successfully represent the continuous fracture progression defined by the Genant scale. 


\subsubsection{Interpretability of generative model} 
In addition to feature extraction, our method also offers a framework for generating both conditional and unconditional samples. Leveraging the generative capability of the model can aid in the interpretability of the decision support system. As illustrated in Fig.~\ref{fig:edited_images_shift}, the images can be manipulated relative to the decision boundary - from healthy vertebra to severe fracture and vice versa. This feature can be utilized to generate counterfactuals by crossing the hyperplane. Fig.~\ref{fig:edited_images_calibrated} demonstrates that this can be extended to visualize if the model is well calibrated to the ordinal regression of Genant grading, by showing the models perception of each grade for a given vertebra. While the last row in Fig.~\ref{fig:edited_images_calibrated} is not well calibrated, the rightmost image augments a barely visible feature. This exaggeration of anatomical changes could guide the radiologist's attention and help estimate the potential progression of pathological changes barely visible in the original image.

\subsubsection{Limitations and future work} 
While the well-calibrated samples in Fig.~\ref{fig:edited_images_calibrated} show impressive visual results, the grading metrics in Tab.~\ref{tab:fracture_results} reveal the limitations of our method. The linear separability of classes in DAE's unsupervised semantic latent space using 2D slices cannot compete with the fully-supervised, end-to-end baseline and 3D methods, motivating an extension to three dimensions.
One reason for this might be a failure to disentangle the fracture of adjacent vertebrae from the centered vertebral body that is classified. This can be observed in the second to last example in Fig.~\ref{fig:edited_images_calibrated}, where the center vertebra remains unchanged while the top one is changing with the severity of the fracture. At the same time, this showcases the explainability of our approach by visualizing the model's understanding of different grades, highlighting its misguided feature attribution. In future work, the network could be explicitly guided to attend to the main vertebrae or a masking applied to the input image. 

\begin{figure}[ht]
  \includegraphics[width=1\linewidth]{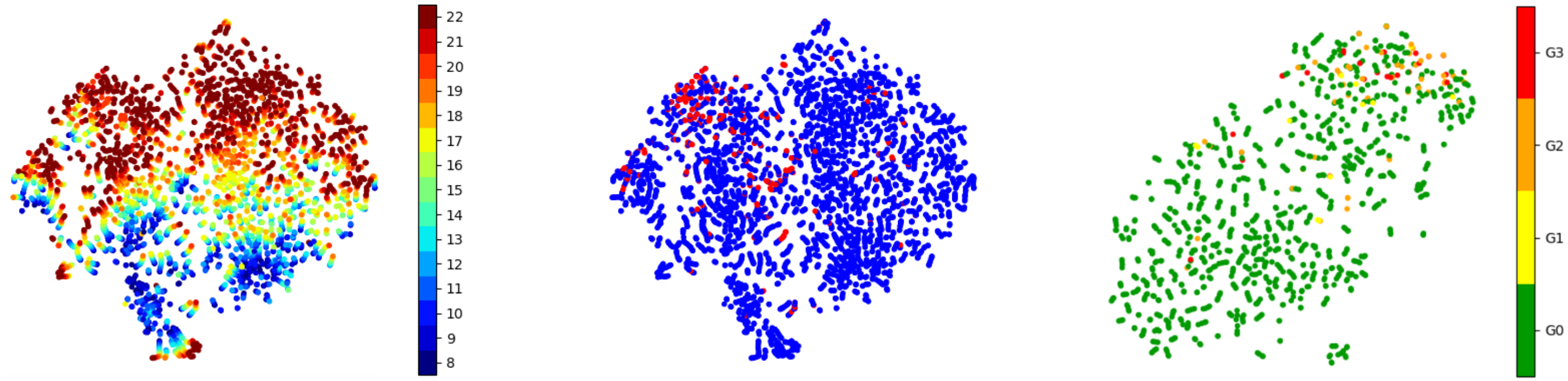}
  \caption{t-SNE projections of $z_{sem}$ of VerSe training set vertebrae encoded by the DAE model. On the left, the vertebra levels from T1 (8) to L5 (22) are visualized. In the center, the same data with healthy vertebrae (blue) and fractures (red) is indicated. On the right, the levels L1-L4 are shown with their Genant grading ranging from G0 to G3.\looseness=-1}
  \label{fig:tsne}
\end{figure}

The t-SNE projection of latents in Fig.~\ref{fig:tsne} shows that the unsupervised generative feature extractor DAE clearly grouped the different vertebral levels ranging from T1 to L5. Since fractures occur more frequently in the lumbar spine, a cluster of fractures can be observed there with many outliers. We hypothesized that training a generative model with all levels can aid the data imbalance since only very few fractures are present in the data per vertebral level. However, the better separatability of Genant grades in the visually similar subset of vertebra L1-L4 (Fig.~\ref{fig:tsne}, right) suggests the investigation of independent models for the different spine segments.

To challenge the assumption of a linear relationship between the distance to the hyperplane and the severity of the fracture, we fitted a polynomial regression to grades G0, G2, and G3. The results in Table \ref{tab:fracture_results} show an improvement over our simple regression model, indicating the presence of a non-linear relationship that warrants further investigation in future research.

\section{Conclusion}
In this work, we deployed a Diffusion Autoencoder (DAE) as an unsupervised, generative feature extractor for the grading of vertebral fractures. Our results demonstrate that a hyperplane can be constructed in the DAE latent space that serves as an effective decision boundary for fracture detection and that the distance from this hyperplane can be leveraged to estimate fracture severity. As we train the feature extractor without any supervision, these results have the potential to scale with more unlabelled data. Notably, our method offers inherent interpretability by enabling clinicians to visualize the inner representations of the model's decision-making process. 
 
\subsubsection{Acknowledgements}
The authors acknowledge the financial support by the Federal Ministry of Education and Research of Germany (BMBF) under project DIVA (FKZ 13GW0469C).

\bibliographystyle{splncs04}
\bibliography{references}
\vspace{1cm}
\hypertarget{supplement}{}
\section*{Supplementary material}
\begin{table}[!htb]
\centering

\caption{Description of the in-house dataset from Klinikum Rechts der Isar and Klinikum der Universität München used for training the diffusion autoencoder.}
\begin{tabular}{|l|l|}
\hline
\textbf{Characteristic}              & \textbf{Value}                                 \\ \hline
Number of patients                   & 465                                           \\ \hline
Median age (years)                   & $\sim$69 ($\pm$12)                             \\ \hline
Scan types                           & Healthy and fractured vertebrae               \\ \hline
Nature of fractures                  & Osteoporotic or malignant                     \\ \hline
Additional features                  & Metallic implants and foreign materials       \\ \hline
CT scanner                        & Heterogeneous              \\ \hline
Scanner setting                 & Heterogeneous              \\ \hline
Field of view & Varying \\ \hline
\end{tabular}
\label{table:dataset_summary}
\end{table}

\begin{table}
\centering

\caption{Dependencies and their respective versions required for implementing the proposed method. The diffusion autoencoder and all neural networks were trained with PyTorch. The SVM classifier and logistic regression were trained with cuML.}
\begin{tabular}{|l|l|}
\hline
\textbf{Library}         & \textbf{Version} \\ \hline
python                    & 3.9.15           \\ \hline
torch                    & 1.8.1           \\ \hline
torchvision              & 0.9.1           \\ \hline
monai                    & 1.0.1           \\ \hline
pytorch-lightning        & 1.4.5           \\ \hline
torchmetrics             & 0.5.0           \\ \hline
scipy                    & 1.5.4           \\ \hline
numpy                    & 1.19.5          \\ \hline
pytorch-fid              & 0.2.0           \\ \hline
lpips                    & 0.1.4           \\ \hline
pandas                   & 1.1.5           \\ \hline
Pillow                   & 8.3.1           \\ \hline
lmdb                     & 1.2.1           \\ \hline
cuml                     &     22.10.1 \\ \hline
scikit-learn              &    1.1.3 \\ \hline
scikit-image               &   0.19.3 \\ \hline
\end{tabular}
\label{table:dependencies}
\end{table}

\begin{table}
\centering

\caption{Hyperparameters used in the diffusion autoencoder training and evaluation. The model was trained on a single Nvidia A40 GPU.}
\begin{tabular}{|l|l|}
\hline
\textbf{Hyperparameter}      & \textbf{Value}         \\ \hline
learning rate                & 0.0001                 \\ \hline
batch\_size                  & 64                     \\ \hline
image\_size                  & 96 x 96                \\ \hline
embedding\_size              & 512 \\ \hline
eval\_every\_samples         & 250,000                \\ \hline
precision                    & half                   \\ \hline
latent\_T\_eval              & 1,000                  \\ \hline
latent\_beta\_scheduler      & linear                 \\ \hline

total\_samples               & 12,000,000             \\ \hline
T                            & 1,000                  \\ \hline
eval\_T                      & 20                      \\ \hline
generation T                 & 100                    \\ \hline
\end{tabular}
\label{table:hyperparameters}
\end{table}

\end{document}